\documentclass[11pt]{article}

\usepackage[preprint]{acl}

\usepackage{times}
\usepackage{latexsym}
\usepackage{microtype}
\usepackage{inconsolata}

\usepackage{xcolor}
\newcommand{\good}[1]{\textcolor{blue}{#1}}
\newcommand{\bad}[1]{\textcolor{red}{#1}}

\usepackage{graphicx}
\usepackage{subcaption}
\usepackage{url}
\usepackage{hyperref}

\usepackage[table]{xcolor}
\usepackage{pifont}

\usepackage{booktabs}
\usepackage{multirow}
\usepackage{makecell}
\usepackage{adjustbox}
\usepackage{float}
\usepackage{tabularx}
\usepackage{array}
\usepackage{caption}
\usepackage[most]{tcolorbox}

\usepackage{arydshln}

\newcolumntype{Y}{>{\raggedright\arraybackslash}X}

\def\method{PASE}

\title{Program-Aided Patchable Skill Evolution}

\author{
    Jiahao Wang$^{1}$\thanks{ \ \ indicates equal contribution.},
    Yuqi Xiong$^{1}$\footnotemark[1],
    Qing Yang$^{1}$,
    Zhipeng Xu$^{2}$, \\
    \textbf{Zhenghao Liu$^{1}$\thanks{ \ \ indicates corresponding author.},
    Zulong Chen$^{2}$,
    Yubin Bao$^{1}$} \\
    $^1$School of Computer Science and Engineering, Northeastern University, Shenyang, China \\
    $^2$Alibaba Group, Hangzhou, China
}

\begin{document}
\raggedbottom

\begingroup
\setlength{\tabcolsep}{0pt}
\maketitle
\endgroup
{
\renewcommand{\thefootnote}{\fnsymbol{footnote}}

\begin{abstract}
Agent skills have emerged as an effective means of enabling LLM agents to solve complex, interactive tasks by encoding reusable knowledge without modifying model parameters. Recent approaches further improve skill adaptation by distilling reusable experience from agent execution trajectories and evolving skills based on downstream task feedback. However, existing methods primarily represent skills as natural language documents, intertwining procedural logic with descriptive knowledge and making skills difficult to maintain and evolve. To address this limitation, we propose \method{} (\textbf{P}rogram-\textbf{A}ided Patchable \textbf{S}kill \textbf{E}volution), a program-aided framework that iteratively evolves skills by associating them with executable programs. Unlike prior approaches that encode trajectory-derived experience primarily in natural language, \method{} separates reusable procedural logic into editable programs, enabling targeted updates to executable components while reducing the need to repeatedly revise natural language skills. Extensive experiments across diverse models and benchmarks demonstrate that \method{} consistently outperforms all baselines, by achieving over 2.8\% improvements. Further analyses show that \method{} exhibits greater evolutionary stability and generalisation capability, highlighting the critical role of programmatic knowledge in skill evolution. All code and data are available at \url{https://github.com/NEUIR/PASE}.

\end{abstract}

\section{Introduction}
\label{sec:introduction}
\begin{figure}[t]
    \centering
    \includegraphics[width=\columnwidth]{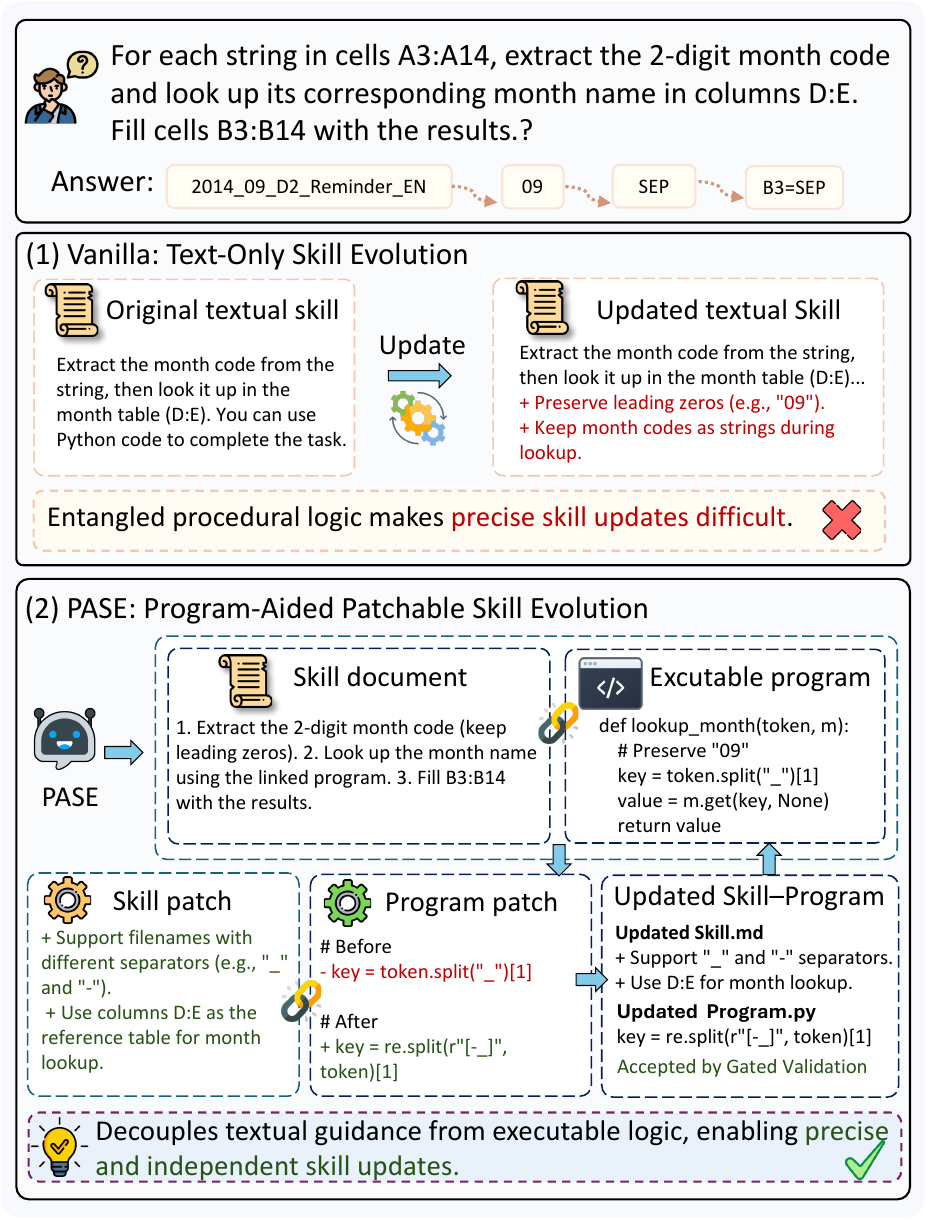}
    \caption{Comparison between vanilla text-only skill-based agents and the PASE framework.}
    \label{fig:pase_intro}
    \vspace{-12pt}
\end{figure}
Leveraging the emergent capabilities of large language models (LLMs), LLM agents have demonstrated strong performance on complex and interactive real-world tasks~\citep{jackson2025aa,merrill2026terminal,jimenez2024swe,eliseeva2025envbench}. By actively interacting with their environments and incorporating environmental feedback, agents can adapt their subsequent reasoning and actions to evolving task conditions~\citep{yao2022react,huang2026envharness,liu2026mitigating}. However, despite their general-purpose capabilities, LLM agents often struggle with domain-specific tasks that require specialized knowledge and operational expertise~\citep{zhang2025survey,ni2026trace2skill,ouyang2026reasoningbank,peng2026mixture,liu2026knowledge}. Effectively addressing such tasks requires reusable domain knowledge, procedural workflows, and operational guidelines that can be applied across task instances.

Agent skills~\citep{anthropic2026skills,zhou2026memento,li2026skillsbench,liu2026well} offer a natural way to equip agents with domain-specific knowledge. Typically represented as natural language documents and provided as context during inference, skills encode reusable knowledge that enables agents to adapt to diverse tasks and environments without fine-tuning~\citep{yang2026skillopt,ni2026trace2skill}. However, conventional skills are often manually authored or generated directly from task descriptions, leaving them largely static after deployment. Without a systematic mechanism for learning from the environment feedbacks, such skills struggle to address recurring failures, accumulate skillable experience from both successful and failed attempts, and adapt to evolving task requirements~\citep{zhou2026comprehensive,xia2026skillrl,ouyang2026reasoningbank,ni2026trace2skill}.

To address this limitation, recent work has explored evolving skills from agent execution trajectories~\citep{alzubi2026evoskill,xiong2026skill,zhang2026coevoskills,wang2026skillgrad,shen2026skillopt,xiong2026lang2act}. These approaches distill reusable experience from execution traces and incorporate the resulting insights into skills through iterative patching. Despite their effectiveness in accumulating experience, they primarily encode the extracted knowledge in natural language, intertwining procedural logic with descriptive knowledge within a single textual artifact. This entanglement makes it difficult to isolate and revise individual behavioral components without inadvertently affecting other parts of a skill, limiting the precision and maintainability of skill patches~\citep{shen2026skillopt,huang2026envharness,gao2023pal}. Consequently, a key challenge is how to enable precise and localized skill updates while retaining the flexibility of natural-language representations.

In this paper, we propose \method{} (\textbf{P}rogram-\textbf{A}ided \textbf{P}atchable \textbf{S}kill \textbf{E}volution), a program-aided framework for iteratively constructing and evolving agent skills. Rather than encoding all procedural details in natural language, \method{} decouples reusable procedural logic from natural-language skill descriptions by extracting recurring executable workflows into standalone programs. This design enables agents to leverage executable procedural knowledge through program execution while preserving the flexibility and readability of natural-language skills. Furthermore, \method{} introduces a patch-based co-evolution mechanism that iteratively updates both skills and their associated programs, exploiting the modularity and maintainability of executable programs to facilitate targeted revisions. By enabling the coordinated evolution of these complementary representations, \method{} supports more precise, maintainable, and effective skill updates.

Our experiments demonstrate the effectiveness and generalizability of \method{}, which consistently outperforms all baselines across benchmarks and different backbone LLMs. By decoupling procedural logic into executable programs, \method{} achieves higher performance and greater stability across evolution rounds than text-only skill evolution, highlighting the importance of programmatic representations for maintaining and evolving procedural knowledge. Further analysis suggests that isolating complex procedural logic in executable programs, rather than encoding it entirely in natural language, facilitates LLMs' understanding and utilization of procedural knowledge during skill evolution.

\section{Related Work}
\label{sec:related_work}

Agent skills are portable artifacts that encapsulate reusable knowledge and procedural workflows, enabling LLM agents to accomplish diverse real-world tasks over extended interactions~\citep{zhou2026comprehensive,anthropic2026skills,jiang2026sok,li2026skillsbench,ma2024spreadsheetbench}. Skills can be inspected, executed, and updated independently of the underlying model, allowing frozen LLMs to adapt to diverse tasks and environments without modifying model parameters~\citep{ni2026trace2skill,yang2026skillopt,wang2026skilltester}. However, conventional skills are often manually authored or generated directly from task-specific descriptions, leaving them largely static after deployment. Without a systematic mechanism for learning from execution experience, such skills struggle to address recurring failures and adapt to evolving task requirements~\citep{shinn2023reflexion,wang2023voyager,liu2026skillrevise,ouyang2026reasoningbank}.

\begin{figure*}[t]
    \centering
    \includegraphics[width=1\linewidth]
    {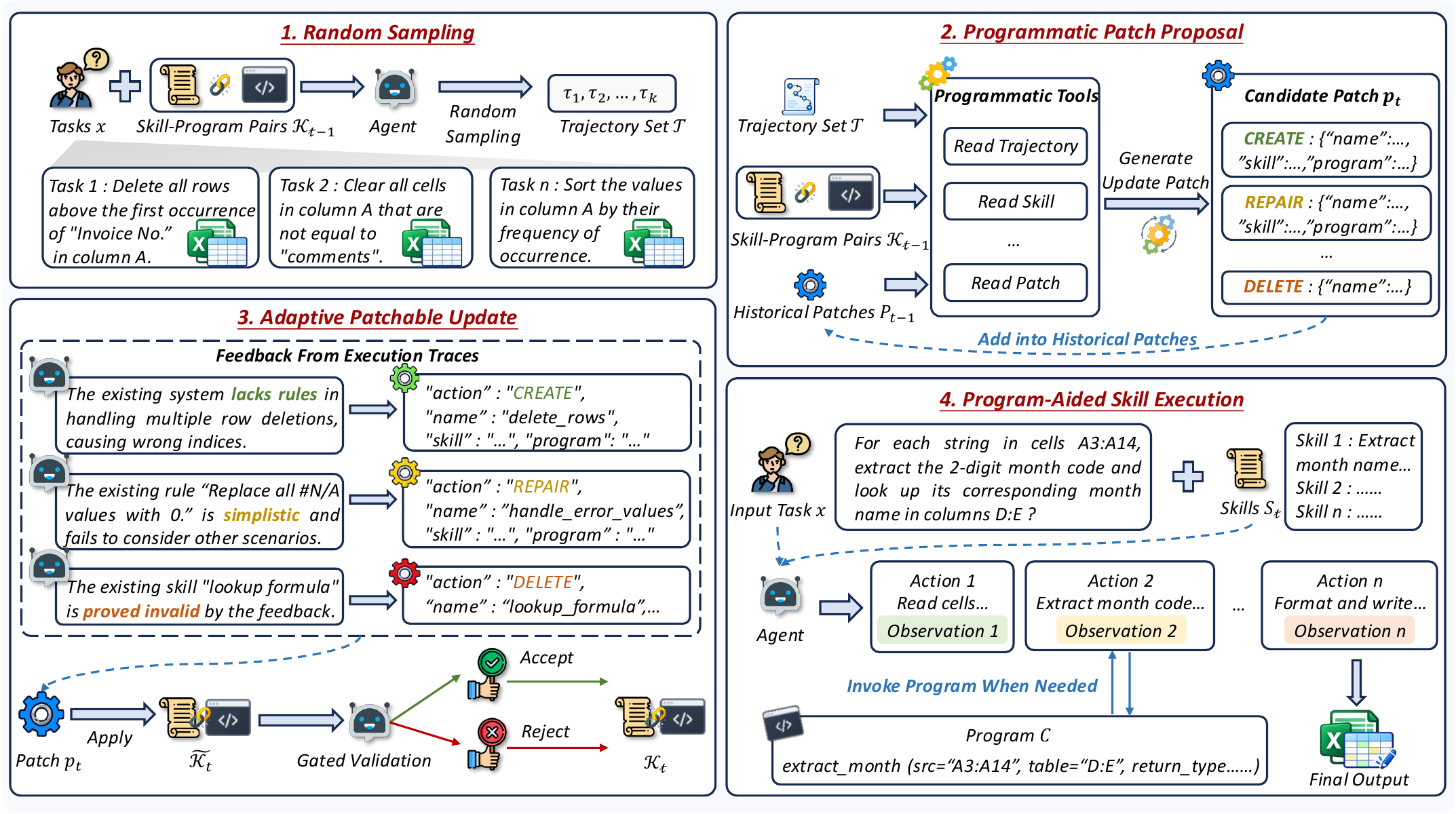}
    \caption{Overview of our \method{} framework.}
    \label{fig:main}
    \vspace{-12pt}
\end{figure*}

To construct and maintain high-quality skills, recent studies have developed specialized frameworks that analyze agent execution trajectories and iteratively evolve skills based on observed outcomes~\citep{ni2026trace2skill,alzubi2026evoskill,agrawal2026gepa,lu2026skill0,ouyang2026skillos}. These methods distill reusable experience from execution traces and translate the extracted insights into skill patches for incremental skill construction and refinement~\citep{alzubi2026evoskill,tang2026wikiskill,ni2026trace2skill}. Another line of work treats skills as optimizable external components and applies training-inspired mechanisms, such as learning-rate control and held-out validation, to improve their quality iteratively~\citep{yang2026skillopt,wang2026skillgrad}. Despite these advances, existing approaches primarily encode reusable knowledge in natural language, intertwining procedural logic with descriptive knowledge within a single textual representation. This entanglement complicates localized modifications and limits the precision and maintainability of skill updates.

Executable programs offer a promissing way to represent procedural knowledge, facilitating LLMs' understanding of complex logic~\citep{wang2024executable,yang2026proceduralskill}. Existing approaches have explored constructing programmatic skills by abstracting procedural logic into reusable callable functions~\citep{wang2023voyager,wang2025inducing,shao2026lifting}. However, representing skills primarily as programs may limit the capture of valuable knowledge that cannot be readily formalized as executable logic, including high-level strategies and descriptive insights derived from successful experiences. In contrast, \method{} does not replace natural-language skills with programs. Instead, it decouples reusable executable workflows from natural-language skill descriptions, preserving diverse forms of procedural knowledge while facilitating more precise and flexible skill evolution.

\section{Methodology}
\label{sec:method}
In this section, we introduce \method{} (\textbf{P}rogram-\textbf{A}ided Patchable \textbf{S}kill \textbf{E}volution), a program-aided iterative framework for constructing and updating agent skills. We first provide the preliminaries of program-aided patchable skill evolution, and then represent the adaptive skill update via programmatic patch proposal mechanism (Sec.~\ref{subsec:3.2}).

\subsection{Preliminaries of Program-Aided Patchable Skill Evolution}
\label{subsec:3.1}

Given an input $x$ and a set of agent skills $\mathcal{S} = \{s_1,s_2,...,s_m\}$, where agent skills encode reusable domain knowledge and task-specific workflows, enabling LLM-based agents to solve specialized tasks by generating reasoning trajectories. To establish the foundation for our method, we first introduce skill-augmented agents and then formalize program-aided skill evolution.

\textbf{Skill-Augmented Agents.} Formally, the agent generates a trajectory $\tau$ for the input $x$ conditioned on the provided skills:
\begin{equation}
\tau
=
\mathcal{M}_\mathrm{infer}(x,\mathcal{S}).
\label{eq:infer}
\end{equation}

To adapt skills to evolving task requirements, existing methods~\citep{alzubi2026evoskill,yang2026skillopt} iteratively refine skills based on agent execution trajectories. Specifically, at evolution step $t$, the framework updates the current skill $\mathcal{S}_t$ from its previous version $\mathcal{S}_{t-1}$ using a batch of training examples $\mathcal{B} = \{(x_i,y_i)\}_{i=1}^k$ and their corresponding execution trajectories:
\begin{equation}
\mathcal{S}_{t}
=
\mathrm{Evolve}\left(
\mathcal{S}_{t-1},
\{(x_i,y_i,\tau_i)\}_{i=1}^k
\right),
\end{equation}
where $x_i$ and $y_i$ denote the input and ground-truth answer of the $i$-th example, respectively, and $\tau_i$ is the corresponding reasoning trajectory generated by the LLM (Eq.~\ref{eq:infer}). However, directly extracting reusable knowledge from these trajectories and encoding it in natural language can entangle procedural logic with descriptive knowledge. Such entanglement makes it difficult to isolate and modify specific behavioral components, limiting the patchability of evolved skills.

\textbf{Program-Aided Skill Evolving.} To address this limitation, \method{} decouples reusable procedural logic from natural-language skills by extracting common executable workflows into standalone programs. Rather than encoding all procedural details in natural language, a skill primarily specifies the task-solving strategy and invokes the corresponding program when needed. Specifically, \method{} identifies core procedural workflows and their supporting operational knowledge from execution trajectories, transforms the workflows into executable programs, and links the resulting programs to the corresponding skills:
\begin{equation}
s_i
\xrightarrow{\mathrm{link}} c_i,
\quad
c_i \in \{\text{code}, \emptyset \},
\end{equation}
where $c_i \in \mathcal{C}$ denotes an executable program. We define $\mathcal{K} = \{(s_i,c_i)\}_{i=1}^m$ as the collection of skill-program pairs, where each skill is associated with its linked program. This representation separates the high-level task-solving strategy from the underlying procedural implementation, allowing individual programs to be independently modified without rewriting the entire skills.

Given an input $x$, the agent then generates its reasoning trajectory through iterative reasoning, action execution, and observation:
\begin{equation}
(r_t,a_t,o_t)
=
\mathcal{M}_\mathrm{infer}
(x,\mathcal{K},\tau_{1:t-1}),
\label{eq:step_infer}
\end{equation}
where $\tau_t=(r_t,a_t,o_t)$ denotes the $t$-th interaction step, and $r_t$, $a_t$, and $o_t$ represent the reasoning, action, and observation, respectively. When the action $a_t$ invokes a linked program $c_j\in\mathcal{C}$, the program is executed in the specified environment to obtain an observation \citep{wang2024executable}:
\begin{equation}
o_t = \mathrm{Exec}(c_j,\mathrm{env}),
\end{equation}
where $\mathrm{env}$ denotes the execution environment. The resulting observation is incorporated into the subsequent reasoning process, enabling the agent to leverage executable procedural knowledge while preserving the flexibility of natural-language skill descriptions. During evolution, skills and programs are jointly updated through a unified framework:
\begin{equation}
\mathcal{K}_{t}
=
\mathrm{Evolve}\left(
\mathcal{K}_{t-1},
\{(x_i,y_i,\tau_i)\}_{i=1}^k
\right).
\label{eq:evolve}
\end{equation}
However, updating these two forms of knowledge requires a reliable and adaptive mechanism to determine which skills or programs to modify and what specific content within them to update.

\subsection{Program-Aided Evolution via Patchable Skill Updating}
\label{subsec:3.2}

To balance efficiency and effectiveness, at each evolution iteration $t$, we randomly sample $k$ tasks with their execution trajectories (Eq.~\ref{eq:evolve}). We include both successful and failed tasks, where successful trajectories provide reusable behavioral patterns, while failed trajectories reveal failure root causes and limitations.

We employ a React-based proposer $\mathcal{M}_{prop}$ to conduct self-reflection across multiple trajectories, distill common experience, and propose a skill update patch $p_t$:
\begin{equation}
p_t =
\mathcal{M}_{\mathrm{prop}}\left(
\mathcal{K}_{t-1},
\mathcal{P}_{t-1},
\{(x_i,y_i,\tau_i)\}_{i=1}^{k}
\right),
\end{equation}
where $\mathcal{P}_{t-1} = \{p_1,p_2, ...,p_{t-1}\}$ represents historical proposed patches before evolution round $t$. The resulting patch is then applied to update the current skills and programs:
\begin{equation}
\widetilde{\mathcal{K}}_t = 
\mathrm{Apply} (p_t, \mathcal{K}_{t-1}),
\label{eq:apply}
\end{equation}
where $\widetilde{\mathcal{K}}_t$ denote the updated skills and programs. They are then evaluated on a held-out validation set, and the acceptance decision is governed by:
\begin{equation}
\mathcal{K}_t
=
\begin{cases}
\widetilde{\mathcal{K}}_t,
&
\mathcal{R}(\widetilde{\mathcal{K}}_t)
>
\mathcal{R}\bigl(\mathcal{K}_{t-1}),
\\[4pt]
\mathcal{K}_{t-1},
&
\text{otherwise},
\end{cases}
\label{gate}
\end{equation}
where $\mathcal{R}(\cdot)$ denotes the validation performance measured on a held-out validation set, which is disjoint from the training set and used exclusively to evaluate skill updates. The update is accepted if it improves validation performance; otherwise, the update at evolution round $t$ is rejected. This mechanism ensures that only empirically beneficial updates are incorporated into the evolving skills, thereby mitigating overfitting to the training set.

\textbf{Patchable Skill Updating.} We represent each patch as an update operation:
\begin{equation}
p = (\mathrm{opt}, s_\text{target}, \mathcal{E}_s, \mathcal{E}_c),
\end{equation}
where $\mathrm{opt} \in \{\mathit{create}, \mathit{repair}, \mathit{delete}\}$ specifies the operation type, and $s_\text{target}$ identifies the target skill. $\mathcal{E}_s$ and $\mathcal{E}_c$ denote the update contents for the skill document and its associated executable program, respectively. The proposer adaptively selects which components to update---the skill document, its associated program, or both---based on the decision of LLM. By decoupling the executable logic from the skill description, our approach enables localized updates to either component without rewriting the entire skill, facilitating more efficient maintenance and evolution of agent skills.

Specifically, these three operations $\mathrm{opt}$ have distinct semantics. The $\mathit{create}$ operation introduces a new skill, optionally with an associated program, where $\mathcal{E}_s$ and $\mathcal{E}_c$ contain the complete content of the new skill document and program, respectively. The $\mathit{delete}$ operation removes the target skill and its associated program, if any, entirely, with both $\mathcal{E}_s$ and $\mathcal{E}_c$ set to $\emptyset$. The $\mathit{repair}$ operation applies fine-grained modifications to an existing skill document and/or its associated program, with $\mathcal{E}_s$ and $\mathcal{E}_c$ specifying sets of line-level edits for the respective components. Each edit identifies a target line and uses the $+$ and $-$ operators to insert new content or delete existing content, respectively. An empty edit set $\emptyset$ indicates that the corresponding component remains unchanged. When a patch contains multiple edits, they are applied sequentially to the current skill and program states according to Eq.~\ref{eq:apply}. 

\textbf{Adaptive Context Reading for Patch Proposal.} The proposer $\mathcal{M}_{prop}$ reads multiple agent execution trajectories, current skills and programs, and historical proposed patches. These files are not pre-loaded into the context but reside in the environment. The proposer retrieves information on demand through predefined programmatic tools, preventing the introduction of irrelevant noise and beyond the model's context window.

We provide \method{} with four tools as follows:

\noindent$\bullet$\ \textbf{read\_trajectory (traj\_id, steps):} Inputs a single trajectory ID and a list of step indices of this trajectory. Returns the raw actions and observations of the specified steps from the trajectory.

\noindent$\bullet$\ \textbf{read\_skill (skill\_name):} Inputs a target skill name. Returns the full content of the corresponding skill document and its linked program.

\noindent$\bullet$\ \textbf{read\_patch (patch\_id):} Inputs a historical patch ID. Returns the full content of the corresponding proposed patch, enabling the proposer to learn from past failures and avoid redundant proposals.

\noindent$\bullet$\ \textbf{finish () :} Submits the final update patch with a reason for the proposal.

The initial user message provides: the task input and the total number of trajectory steps; the name of skills; and the ids, operations and target name of updated skills of all historical patches.

The proposer first reads multiple trajectories to distill common experience, then retrieve relevant historical proposed patches to refine its experience, and inspect relevant skills to translate the refined experience into updates to the current skills. After retrieve enough information, the proposer finally propose a patch $p_t$ with \texttt{finish}().

\begin{table*}[t]
\begin{center}
\resizebox{\textwidth}{!}{
\begin{tabular}{l ccc cccccc cc c}
\hline

\multirow{3}{*}{\textbf{Methods}}
& \multicolumn{3}{c}{\textbf{In Domain}}
& \multicolumn{8}{c}{\textbf{Out of Domain}}
& \multirow{3}{*}{\textbf{Avg.↑}}
\\

\cmidrule(lr){2-4}
\cmidrule(lr){5-12}
&
\multicolumn{3}{c}{\textbf{SpreadsheetBench}}
& \multicolumn{6}{c}{\textbf{Domain-Spreadsheet}}
& \multicolumn{2}{c}{\textbf{TableQA}}
&
\\
\cmidrule(lr){2-4}
\cmidrule(lr){5-10}
\cmidrule(lr){11-12}

&
\textbf{Vrf↑}
& \textbf{Soft↑}
& \textbf{Hard↑}
& \textbf{FIN↑}
& \textbf{HR↑}
& \textbf{RE↑}
& \textbf{SAL↑}
& \textbf{SC↑}
& \textbf{Overall↑}
& \textbf{WikiTQ↑}
& \textbf{HiTab↑}
&
\\

\hline

\rowcolor{cyan!8}
\multicolumn{13}{c}{\textit{Qwen3.6-27B}}
\\
\hline

No Skill
& 43.81 & 19.57 & 20.76
& 26.17 & 18.02 & 23.97 & 10.34 & 24.13 & 21.07
& 51.35 & 42.63 & 32.04
\\

Human Skill
& 64.76 & 42.67 & 25.58
& 53.09 & 41.80 & 43.45 & 38.70 & 42.36 & 44.02
& 70.23 & 46.90 & 48.97
\\

LLM Skill
& 43.69 & 31.75 & 17.69
& 28.64 & 30.09 & 33.33 & 27.20 & 37.75 & 31.91
& 83.45 & 67.60 & 46.16
\\

Trace2Skill
& 65.60 & 46.59 & 29.09
& 51.11 & 47.39 & 45.32 & \textbf{40.61} & 49.17 & 47.46
& 83.45 & 69.64 & 57.03
\\

EvoSkill
& 68.10 & 49.15 & 31.29
& 48.40 & 48.11 & 45.32 & 35.25 & 51.01 & 46.92
& 84.60 & 72.18 & 58.27
\\

\textbf{\method{}}
& \textbf{82.98} & \textbf{53.07} & \textbf{37.57}
& \textbf{54.32} & \textbf{50.45} & \textbf{51.69}
& 39.85 & \textbf{53.22} & \textbf{50.76}
& \textbf{85.39} & \textbf{74.10} & \textbf{62.79}
\\

\hline

\rowcolor{cyan!8}
\multicolumn{13}{c}{\textit{Qwen3.7-Plus}}
\\
\hline

No Skill
& 51.07 & 35.65 & 22.66
& 19.75 & 24.68 & 26.59 & 22.22 & 25.23 & 23.78
& 84.52 & 65.92 & 45.15
\\

Human Skill
& 67.02 & 45.83 & 29.53
& 54.81 & 43.60 & 43.82 & 36.40 & 39.23 & 43.77
& 85.00 & 62.95 & 55.07
\\

LLM Skill
& 63.45 & 42.23 & 27.05
& 33.09 & 38.92 & 38.95 & 26.44 & 39.59 & 36.34
& 84.64 & 68.03 & 52.31
\\

Trace2Skill
& 69.88 & 47.05 & 30.56
& 52.10 & 50.09 & 47.94 & 43.30 & 53.22 & 50.17
& 84.80 & 72.24 & 59.28
\\

EvoSkill
& 78.93 & 52.83 & 36.70
& 55.31 & 52.07 & 53.56 & 42.15 & 54.51 & 52.29
& \textbf{86.54} & 74.29 & 62.95
\\

\textbf{\method{}}
& \textbf{86.67} & \textbf{54.56} & \textbf{39.91}
& \textbf{63.46} & \textbf{56.58} & \textbf{58.80}
& \textbf{43.68} & \textbf{58.93} & \textbf{57.21}
& 86.14 & \textbf{76.21} & \textbf{66.26}
\\

\hline

\rowcolor{cyan!8}
\multicolumn{13}{c}{\textit{Gemma4-31B}}
\\
\hline

No Skill
& 49.05 & 34.67 & 21.20
& 34.81 & 42.34 & 39.33 & 32.95 & 39.96 & 38.60
& 83.06 & 63.01 & 48.87
\\

Human Skill
& 65.00 & 41.40 & 28.51
& \textbf{49.88} & 36.22 & 31.84 & 32.57 & 36.65 & 38.01
& 80.29 & 59.73 & 51.00
\\

LLM Skill
& 44.05 & 30.85 & 20.61
& 26.17 & 34.77 & 35.21 & 29.12 & 29.28 & 30.92
& \textbf{84.36} & 61.59 & 45.24
\\

Trace2Skill
& 59.05 & 38.69 & 27.63
& 44.20 & 43.96 & 40.82 & 35.63 & 45.12 & 42.84
& 83.53 & 65.86 & 53.11
\\

EvoSkill
& 70.83 & 44.81 & 29.09
& 46.91 & 46.31 & 43.82 & \textbf{37.16} & 47.70 & 45.30
& 82.19 & 68.96 & 56.37
\\

\textbf{\method{}}
& \textbf{73.21} & \textbf{48.59} & \textbf{35.67}
& 49.63 & \textbf{49.19} & \textbf{46.82}
& \textbf{37.16} & \textbf{49.72} & \textbf{47.56}
& 81.43 & \textbf{73.30} & \textbf{59.14}
\\

\hline
\end{tabular}
}
\caption{
Overall performance.
}
\label{tab:main_results}
\vspace{-12pt}
\end{center}
\end{table*}

\section{Experimental Methodology}
\label{sec:experiments}
This section describes the datasets, evaluation metric, and baselines, followed by implementation details of our experiments.

\textbf{Datasets.} Our experiments is first evaluated across two spreadsheet benchmarks: Spreadsheetbench \citep{ma2024spreadsheetbench} and Domain-Spreadsheet \citep{chi2026spreadsheet}, where agents manipulate xlsx files through executable Python code. Additionally, following prior work \citep{ni2026trace2skill}, we also report results on two TableQA benchmarks: WikiTableQuestions \citep{pasupat2015compositional} and HiTab \citep{cheng2022hitab}. We convert their input tables into spreadsheet format and solve the tasks through spreadsheet manipulation. More details are provided in Appendix~\ref{appendix: dataset}.

\textbf{Baselines.} We compare \method{} with several representative baseline methods. The No Skill baseline directly prompts the inference agent to solve tasks using its internal knowledge. The Human Skill adopts Anthropic's official XLSX skill, an expert-curated resource providing best practices and procedural guidance for spreadsheet manipulation. The LLM Skill is generated by an LLM based on task descriptions and its internal knowledge, without leveraging execution trajectories. Then we compare our method against two representative skill-evolution baselines. Trace2Skill \citep{ni2026trace2skill} distills trajectory-local lessons from agent executions and hierarchically consolidates them into transferable skills. EvoSkill \citep{alzubi2026evoskill} iteratively extracts and refines reusable skills from execution experience. For a fair comparison, all methods employ the same backbone LLM and infer prompt template in our experiments.

\textbf{Evaluation Metric.} Following the prior work \citep{ni2026trace2skill}, we report Soft, Hard, and Verified accuracy for SpreadsheetBench, denotation accuracy for WikiTableQuestions, and execution accuracy for HiTab. For Domain-Spreadsheet, we evaluate \method{} across five domains, including advanced finance (FIN), human resources (HR), real estate (RE), sales (SAL), and supply chain (SC). We follow the official evaluation protocol \citep{chi2026spreadsheet} and report Pass@1. These metrics measure whether the generated spreadsheets or answers satisfy the corresponding ground-truth criteria.

\textbf{Implementation Details.} We experiment with both open-weight and closed models to evaluate our method and baselines. For open-weight models, we select two models from different model families, Qwen3.6-27B \citep{qwen2026qwen36_27b} and Gemma-4-31B-it \citep{team2026gemma}. For closed models, we use Qwen3.7-Plus \citep{qwen2026qwen37plus}. For skill self-evolution, we use SpreadsheetBench-Verified and split its 400 samples into 80 training examples, 40 held-out validation examples, and the remaining 280 examples for final performance evaluation and reporting. All other datasets are used exclusively for evaluation to assess the out-of-domain generalization of the evolved skills.

\begin{table*}[t]
\begin{center}
\begin{tabular}{l c cc c}
\hline

\multirow{2}{*}{\textbf{Methods}}
& \multicolumn{1}{c}{\textbf{In Domain}}
& \multicolumn{2}{c}{\textbf{Out of Domain}}
& \multirow{2}{*}{\textbf{Avg.↑}}
\\
\cmidrule(lr){2-2}
\cmidrule(lr){3-4}

&
\textbf{SpreadsheetBench}
& \textbf{Domain-Spreadsheet}
& \textbf{TableQA}
&
\\

\hline

Vanilla LLM
& 43.81
& 21.07
& 42.63
& 35.84
\\

w/ Text Skill
& 78.93
& 49.68
& 71.31
& 66.64
\\

w/ Programmatic Skill
& 80.71
& 50.17
& 72.61
& 67.83
\\
\hline

\method{}
& \textbf{82.98}
& \textbf{50.76}
& \textbf{74.10}
& \textbf{69.28}
\\

w/o Program Evolution
& 77.26
& 46.73
& 69.64
& 64.54
\\

w/o Adaptive Reading
& 78.10
& 47.46
& 70.51
& 65.36
\\

w/o Skill Updating
& 77.74
& 47.56
& 69.76
& 65.02
\\

\hline

\end{tabular}
\caption{
Ablation study of \method{}.
}
\label{tab:ablation_results}
\vspace{-12pt}
\end{center}
\end{table*}

\section{Evaluation Results}
\label{sec:results}
In this section, we first evaluate the overall performance of \method{} and conduct ablation studies to assess the contribution of each component. We then analyze how the cross-model transferability of \method{} evolves across different skill creators and skill users. Finally, we present a case study in Appendix~\ref{appendix:case_study}.

\subsection{Overall Performance}
\label{5.1}
Table~\ref{tab:main_results} summarizes the performance of \method{} across different models and benchmarks. Overall, \method{} consistently outperforms the baselines across different backbone LLMs, demonstrating its effectiveness in enhancing LLM's capability.

Compared with the No Skill baseline, incorporating external knowledge through agent skills consistently improves performance across different tasks. Specifically, Human Skill yields moderate improvements across benchmarks, demonstrating the value of expert-crafted knowledge. In contrast, LLM Skill performs comparably to the No Skill baseline and even underperforms it on Gemma-4-31B, highlighting the limitations of LLM-generated skills. Existing skill evolution methods further improve the performance of LLM-generated skills by distilling reusable experience from agent execution trajectories, even surpassing Human Skill. Building on these advances, \method{} achieves further improvements over existing skill evolution methods on both in-domain and out-of-domain benchmarks, demonstrating the benefits of decoupling skill descriptions from their associated executable programs.
Specifically, \method{} achieves a performance improvement of over 14\% on the in-domain SpreadsheetBench-Verified benchmark, demonstrating its effectiveness in acquiring and refining domain-specific skills. More importantly, \method{} generalizes well to out-of-domain tasks, achieving the best performance across all evaluated out-of-domain benchmarks, with improvements of over 3.4\% on Domain-Spreadsheet and 2.8\% on TableQA. These consistent gains suggest that \method{} captures reusable knowledge that extends beyond the training tasks, rather than merely overfitting.

\subsection{Ablation Study}
\label{5.2}

To investigate the contribution of each component in \method{}, we conduct an ablation study using Qwen3.6-27B.

As shown in Table~\ref{tab:ablation_results}, we first investigate the effectiveness of different skill representations. Specifically, Vanilla LLM solves tasks without external skill guidance; \textit{w/ Text Skill} represents each skill entirely as a natural-language document; and \textit{w/ Programmatic Skill} represents each skill exclusively as an executable program. The results show that both textual and programmatic skills substantially outperform Vanilla LLM, demonstrating the importance of external knowledge in guiding LLM-based agents. Moreover, the \textit{w/ Programmatic Skill} model achieves higher performance, suggesting that executable programs can provide more precise and actionable procedural guidance. Furthermore, \method{} achieves the best overall performance, highlighting the complementary roles of natural-language skills and executable programs.

We further investigate the contributions of three key components through ablation studies: (1) \method{} w/o Program Evolution, which evolves only textual skills; (2) \method{} w/o Adaptive Reading, which directly feeds all execution trajectories, skills, and historical proposals into the model context; and (3) \method{} w/o Skill Updating, which disables the evolution of both textual skills and their associated programs. As shown by the results, removing any of these components leads to performance degradation. In particular, the substantial drop caused by removing program evolution highlights the importance of continuously refining executable procedural knowledge beyond textual skill descriptions alone. The performance drop without adaptive reading highlights the importance of retrieving relevant information from execution trajectories on demand when evolving skills and programs, rather than indiscriminately incorporating all available context. Meanwhile, disabling skill updating confirms the importance of iteratively refining both skill documents and their associated programs. Overall, these results validate the complementary roles of procedural knowledge evolution, adaptive information access, and joint skill updating in \method{}.

\begin{figure}[t]
\centering
\begin{subfigure}[t]{0.48\linewidth}
    \centering
    \includegraphics[width=\linewidth]{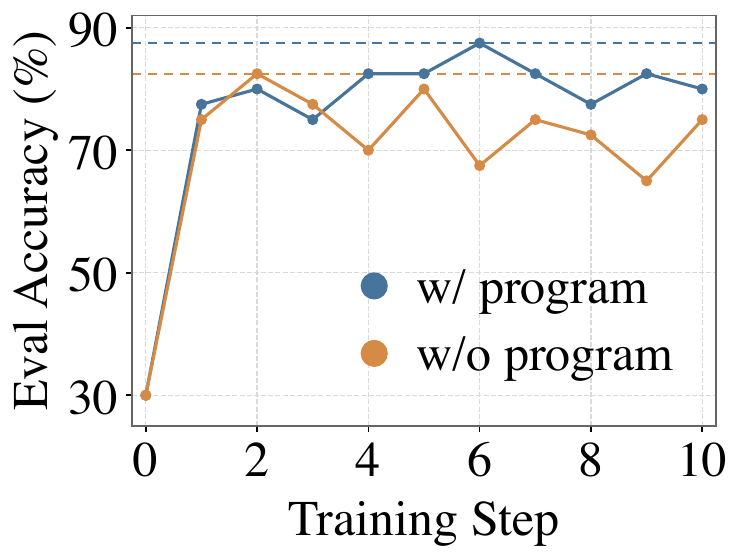}
    \caption{Evaluation accuracy across evolution steps.}
    \label{fig:character:eval}
\end{subfigure}%
\hfill
\begin{subfigure}[t]{0.44\linewidth}
    \centering
    \includegraphics[width=\linewidth]{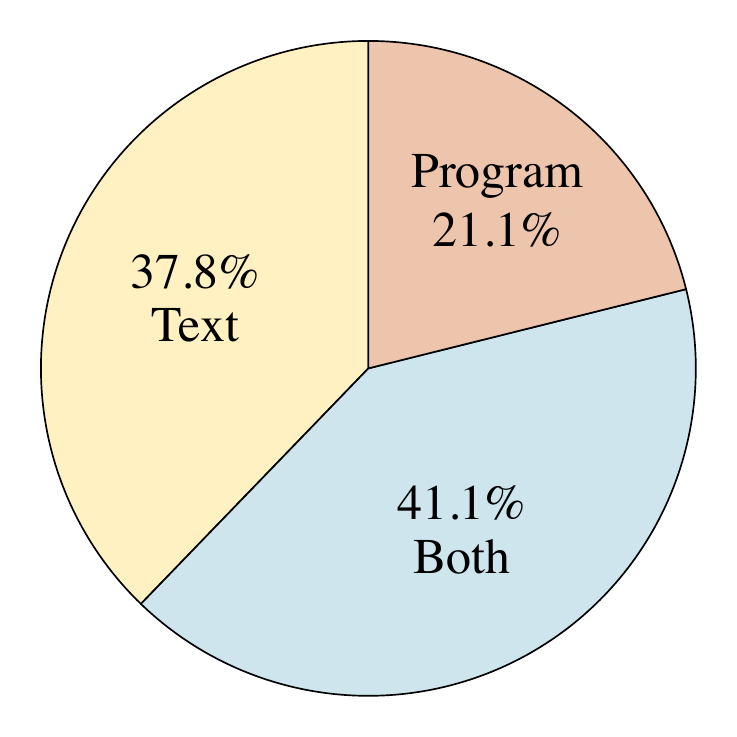}
    \caption{The category distribution of patchable update.}
    \label{fig:character:update}
\end{subfigure}%
\hfill
\begin{subfigure}[t]{0.44\linewidth}
    \centering
    \includegraphics[width=\linewidth]{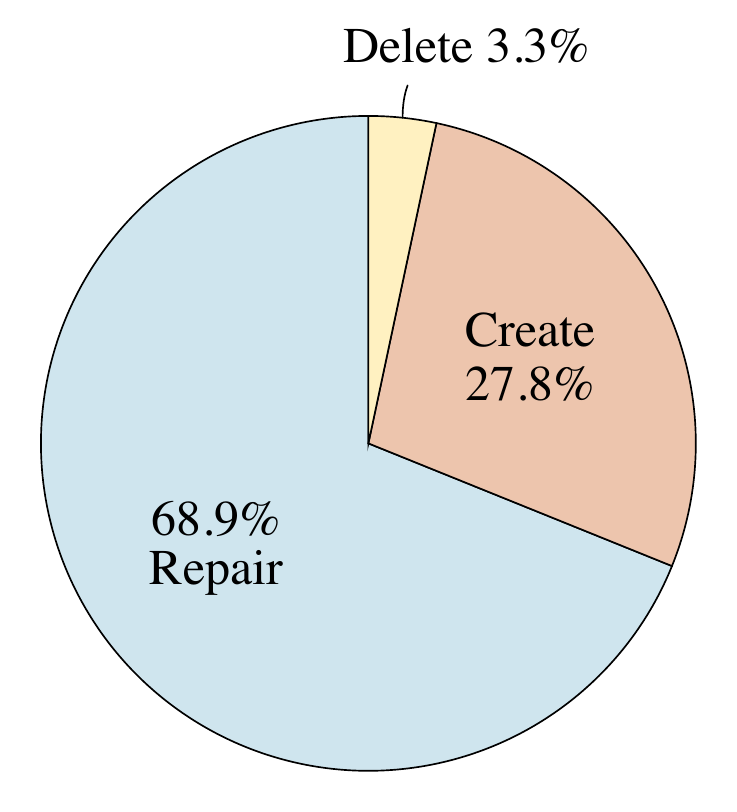}
    \caption{The operation distribution of patch.}
    \label{fig:character:opration}
\end{subfigure}%
\hfill
\begin{subfigure}[t]{0.48\linewidth}
    \centering
    \includegraphics[width=\linewidth]{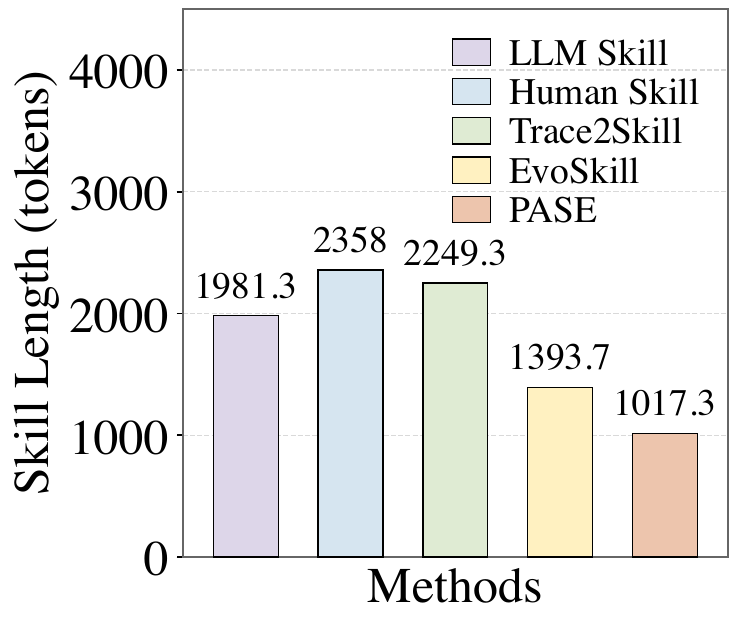}
    \caption{The lengths of textual skill across different methods.}
    \label{fig:character:length}
\end{subfigure}

\caption{Skill evolution characteristics of \method{}. Figure~\ref{fig:character:eval} analyzes the statistics of skill updates, while Figures~\ref{fig:character:update} and~\ref{fig:character:opration} characterize the evolution and operation of the update mechanism. Figure~\ref{fig:character:length} compares textual skill lengths, illustrating how executable programs help reduce the textual burden of skill evolution.}
\label{fig:character}
\end{figure}

\subsection{Skill Evolution Characteristics in \method{}}
\label{5.3}
As shown in Figure~\ref{fig:character}, we analyze the characteristics of skill evolution in \method{} from three perspectives: performance progression, patch operations, and the compactness of evolved skill documents.

In Figure~\ref{fig:character:eval}, we first investigate the role of programs in skill evolution by tracking evaluation accuracy across evolution steps. Specifically, \textit{w/ program} denotes the full \method{} setting, whereas \textit{w/o program} removes skill-associated programs while keeping all other settings unchanged. The results show that \method{} progressively improves its evaluation performance through iterative updates with gated validation, reaching its peak at the sixth evolution step. Moreover, it exhibits smaller performance fluctuations and lower variance across evolution steps, indicating more stable skill evolution. In contrast, the \textit{w/o program} variant peaks at the second step, while subsequent updates introduce greater performance fluctuations without sustained improvements. 

Furthermore, Figures~\ref{fig:character:update} and~\ref{fig:character:opration} illustrate the distributions of updated components and patch operations during skill evolution, respectively. Notably, programs are involved in 62.2\% of all updates, including 21.1\% that update programs alone and 41.1\% that update both programs and skill documents. This high proportion highlights the central role of executable programs in skill evolution. Meanwhile, the patch proposer tends to refine existing skills rather than delete them outright, adaptively selecting whether to update skill documents, their associated programs, or both. These findings demonstrate the flexibility of our patch-based mechanism in supporting fine-grained skill refinement through independently manageable programs. Finally, we further compare the lengths of the final skill documents generated by different methods in Figure~\ref{fig:character:length}. The results show that \method{} produces shorter and more concise skill documents by encapsulating core procedural logic in standalone executable programs, thereby reducing redundant textual content.

\begin{figure}[t]
\centering

\begin{subfigure}[t]{0.48\linewidth}
    \centering
    \includegraphics[width=\linewidth]{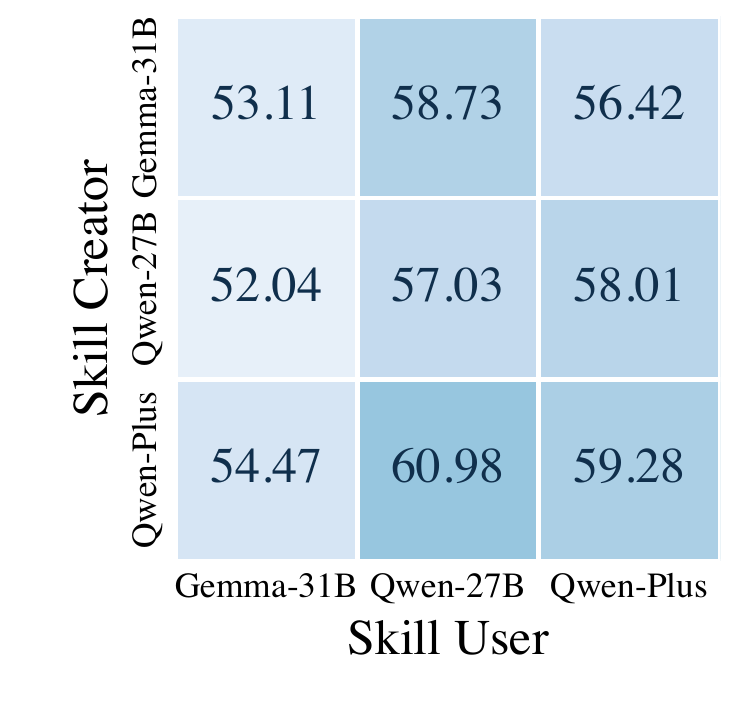}
    \caption{Trace2Skill.}
    \label{fig:transfer:c}
\end{subfigure}%
\hfill
\begin{subfigure}[t]{0.48\linewidth}
    \centering
    \includegraphics[width=\linewidth]{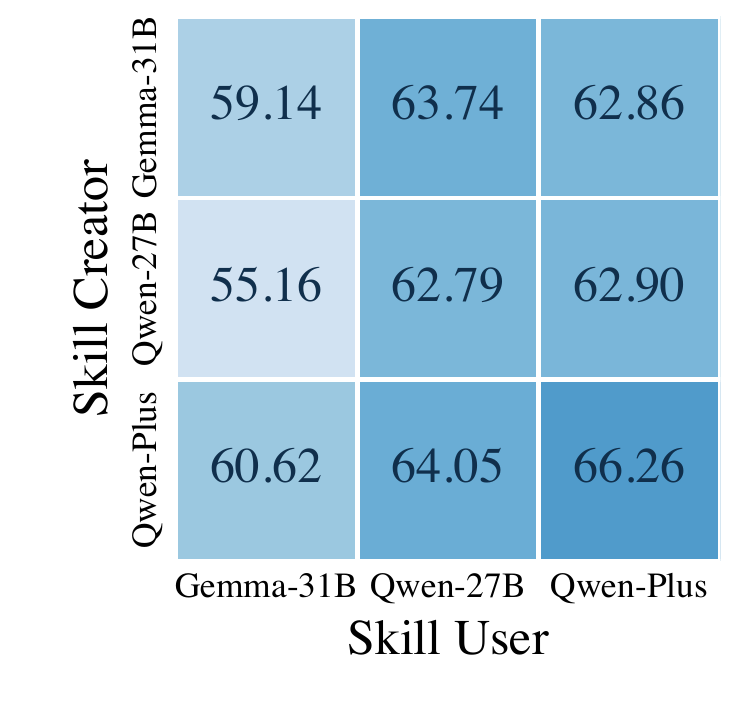}
    \caption{\method{}.}
    \label{fig:transfer:d}
\end{subfigure}

\caption{Generalization of skills learned by different skill evolution methods.}
\label{fig:transfer}
\vspace{-12pt}
\end{figure}
\subsection{Generalization of Evolved Skills in \method{}}
To comprehensively evaluate the generalization of skills evolved by \method{}, we investigate their transferability across different skill creators and skill users, as shown in Figure~\ref{fig:transfer}.

As shown in Figure~\ref{fig:transfer}, \method{} consistently outperforms Trace2Skill across all nine skill creator-user combinations, with improvements of at least 3.07\%. These consistent gains indicate that the effectiveness of \method{} does not depend on specific creator-user combinations. By decoupling executable programs from textual skill descriptions, \method{} facilitates the transfer of procedural knowledge across models and enables diverse LLMs with varying capabilities to benefit from evolved skills.

Then, we further evaluate the transferability of skills across models with different capacities and architectures. Within the Qwen family, transferring skills from Qwen3.6-27B to Qwen3.7-Plus yields a 4.89\% improvement over Trace2Skill. The advantage also extends across model architectures: transferring skills from Gemma-4-31B to Qwen3.6-27B achieves a gain of 5.01\% over Trace2Skill. These results demonstrate that skills evolved by \method{} remain effective across different model families without additional adaptation. By separating descriptive knowledge from executable procedural logic, \method{} reduces reliance on model-specific natural language instructions and facilitates the reuse of executable programs across heterogeneous LLMs.

\section{Conclusion}
\label{sec:conclusion}
This paper proposes \method{} (\textbf{P}rogram-\textbf{A}ided Agent \textbf{S}kill \textbf{E}volution), a novel skill evolution framework that leverages executable programs to decouple core procedural logic from supporting knowledge in natural-language skills. Through patch-based proposals, \method{} adaptively evolves both skills and their associated programs, enabling more precise skill construction and refinement. Extensive experiments demonstrate that \method{} consistently outperforms existing skill evolution methods and produces skills with strong generalization and transferability, improving performance on long-horizon interactive tasks.

\section*{Limitations}
\label{sec:limitation}

Although our approach demonstrates convincing performance in guiding LLMs to evlove skills effectively, it has some limitations. First, following prior work, we directly inject active skills into the agent's prompt to isolate skill quality from potential confounding effects of skill retrieval. Consequently, our evaluation does not account for skill retrieval, which becomes increasingly important as the skill repository grows. Second, our programs rely on explicit invocation by the inference agent. However, a ReAct-based agent equipped with a Bash tool may bypass predefined programs and implement the underlying logic independently, establishing a reliable mechanism to ensure program execution remains a persistent challenge. Finally, our validation gating accepts only proposals that improve the validation score, potentially excluding neutral updates that maintain immediate performance but could facilitate future improvements. We adopt this criterion following prior evolution framework \citep{alzubi2026evoskill} to ensure a fair comparison.


\bibliography{custom}

\appendix

\clearpage
\appendix

\section{Appendix}

\subsection{License}

All datasets used in this work are publicly available under licenses that permit academic research and noncommercial use, including SpreadsheetBench, Domain-Spreadsheet, and WikiTableQuestions (CC BY-SA 4.0), and HiTab (Computational Use of Data Agreement v1.0).

\subsection{Additional Dataset Details}
\label{appendix: dataset}

We evaluate \method{} on one in-domain benchmark, SpreadsheetBench \citep{ma2024spreadsheetbench}, and three out-of-domain benchmarks: Domain-Spreadsheet \citep{chi2026spreadsheet}, WikiTableQuestions \citep{pasupat2015compositional}, and HiTab \citep{cheng2022hitab}. Detailed dataset statistics are presented in Table~\ref{tab:dataset_examples}. Following prior work \citep{ni2026trace2skill}, we adapt the two TableQA benchmarks into spreadsheet manipulation tasks. Specifically, we convert the original JSON-formatted tables into XLSX spreadsheets and instruct the agent to write its predicted answer into fixed cell A1.

\textbf{SpreadsheetBench.}
SpreadsheetBench \citep{ma2024spreadsheetbench} is a real-world spreadsheet manipulation benchmark comprising 912 tasks collected from online Excel forums. It covers diverse operations, including data extraction, formula calculation, cell modification, and formatting. We evaluate on both the full benchmark (912 tasks) and SpreadsheetBench-Verified, an expert-verified subset containing 400 tasks.

\textbf{Domain-Spreadsheet.}
Domain-Spreadsheet \citep{chi2026spreadsheet} evaluates spreadsheet manipulation capabilities across professional domains, including finance, human resources, real estate, sales, and supply chain management. These tasks involve domain-specific knowledge and complex analytical workflows. We select five subsets, including advanced finance (FIN), human resources (HR), real estate (RE), sales (SAL), and supply chain (SC), comprising 677 tasks in total.
\begin{table}[t]
\centering
\small
\begin{tabular}{lr}
\toprule
\textbf{Dataset} & \textbf{\# Examples} \\
\midrule
SpreadsheetBench & 912 \\
\quad\quad Verified & 400 \\
\quad\quad Full & 912 \\
\midrule
Domain-Spreadsheet & 677 \\
\quad\quad FIN & 135 \\
\quad\quad HR & 185 \\
\quad\quad RE & 89 \\
\quad\quad SAL & 87 \\
\quad\quad SC & 181 \\
\midrule
WikiTableQuestions & 1,000 \\
HiTab & 1,000 \\
\bottomrule
\end{tabular}
\caption{Statistics of the evaluation datasets.}
\label{tab:dataset_examples}
\vspace{-12pt}
\end{table}

\textbf{WikiTableQuestions.}
WikiTQ \citep{pasupat2015compositional} is a table question-answering benchmark consisting of natural-language questions grounded in semi-structured Wikipedia tables. It emphasizes compositional reasoning over diverse topics, requiring operations such as filtering, comparison, and aggregation. We sample 1,000 examples from the official test set for evaluation.

\textbf{HiTab.}
HiTab \citep{cheng2022hitab} is a hierarchical table question-answering benchmark constructed from statistical reports and Wikipedia tables. Unlike conventional flat-table benchmarks, HiTab requires reasoning over multi-level row and column hierarchies, as well as complex numerical relationships. We sample 1,000 examples from its test set for evaluation.

\begin{table*}[t]
\begin{center}
\resizebox{\textwidth}{!}{
\begin{tabular}{l ccc cccccc cc c}
\hline

\multirow{3}{*}{\textbf{Skill Creator}}
& \multicolumn{3}{c}{\textbf{In Domain}}
& \multicolumn{8}{c}{\textbf{Out of Domain}}
& \multirow{3}{*}{\textbf{Avg.↑}}
\\
\cmidrule(lr){2-4}
\cmidrule(lr){5-12}

&
\multicolumn{3}{c}{\textbf{SpreadsheetBench}}
& \multicolumn{6}{c}{\textbf{Domain-Spreadsheet}}
& \multicolumn{2}{c}{\textbf{TableQA}}
&
\\
\cmidrule(lr){2-4}
\cmidrule(lr){5-10}
\cmidrule(lr){11-12}
&
\textbf{Vrf↑}
& \textbf{Soft↑}
& \textbf{Hard↑}
& \textbf{FIN↑}
& \textbf{HR↑}
& \textbf{RE↑}
& \textbf{SAL↑}
& \textbf{SC↑}
& \textbf{Overall↑}
& \textbf{WikiTQ↑}
& \textbf{HiTab↑}
&
\\

\hline

\rowcolor{cyan!8}
\multicolumn{13}{c}{\textit{Skill User: Qwen3.6-27B}}
\\
\hline

No Skill
& 43.81 & 19.57 & 20.76
& 26.17 & 18.02 & 23.97 & 10.34 & 24.13 & 21.07
& 51.35 & 42.63 & 32.04
\\

Qwen3.6-27B
& \textbf{82.98} & 53.07 & \textbf{37.57}
& 54.32 & 50.45 & 51.69 & 39.85 & 53.22 & 50.76
& 85.39 & 74.10 & 62.79
\\

Gemma4-31B
& 76.90 & 51.66 & 35.75
& \textbf{62.47} & 52.79 & 49.81 & \textbf{41.38}
& \textbf{60.41} & 54.90
& 85.31 & \textbf{77.76} & 63.74
\\

Qwen3.7-Plus
& 82.74 & \textbf{54.29} & 37.50
& 60.49 & \textbf{55.86} & \textbf{55.06} & 39.46
& 57.27 & \textbf{54.95}
& \textbf{85.47} & 72.55 & \textbf{64.05}
\\

\hline

\rowcolor{cyan!8}
\multicolumn{13}{c}{\textit{Skill User: Qwen3.7-Plus}}
\\
\hline

No Skill
& 51.07 & 35.65 & 22.66
& 19.75 & 24.68 & 26.59 & 22.22 & 25.23 & 23.78
& 84.52 & 65.92 & 45.15
\\

Qwen3.6-27B
& 81.90 & 53.83 & 37.94
& 54.81 & 49.91 & 51.31 & 39.08 & 52.67 & 50.42
& 85.43 & 75.34 & 62.90
\\

Gemma4-31B
& 78.57 & 52.31 & 36.40
& 55.80 & 52.79 & 49.44 & 38.31 & 51.75 & 50.81
& 83.57 & \textbf{80.42} & 62.86
\\

Qwen3.7-Plus
& \textbf{86.67} & \textbf{54.56} & \textbf{39.91}
& \textbf{63.46} & \textbf{56.58} & \textbf{58.80}
& \textbf{43.68} & \textbf{58.93} & \textbf{57.21}
& \textbf{86.14} & 76.21 & \textbf{66.26}
\\

\hline

\rowcolor{cyan!8}
\multicolumn{13}{c}{\textit{Skill User: Gemma4-31B}}
\\
\hline

No Skill
& 49.05 & 34.67 & 21.20
& 34.81 & 42.34 & 39.33 & 32.95 & 39.96 & 38.60
& 83.06 & 63.01 & 48.87
\\

Qwen3.6-27B
& 64.40 & 42.23 & 29.39
& 53.09 & 45.59 & 44.19 & 33.33 & 44.01 & 44.90
& 83.41 & 67.04 & 55.16
\\

Gemma4-31B
& 73.21 & 48.59 & \textbf{35.67}
& 49.63 & 49.19 & 46.82 & \textbf{37.16} & 49.72 & 47.56
& 81.43 & \textbf{73.30} & 59.14
\\

Qwen3.7-Plus
& \textbf{77.38} & \textbf{49.22} & 34.87
& \textbf{58.02} & \textbf{51.71} & \textbf{47.19}
& 36.02 & \textbf{50.46} & \textbf{50.02}
& \textbf{84.60} & 71.44 & \textbf{60.62}
\\

\hline
\end{tabular}
}
\caption{
Cross-model skill transfer results.
}
\label{tab:cross_model_transfer}
\end{center}
\vspace{-12pt}

\end{table*}

\subsection{Additional Experimental Details}

For all the baselines, we utilize the same generation settings. We utilize the vllm framework to deploy open-weight LLMs and Qwen api for close-weight LLMs. During inference, we set the temperature to 0.6, the maximum context length to 131,072 tokens, and the maximum number of ReAct iterations to 30, without explicitly configuring a thinking mode. During evolution, we increase the maximum number of ReAct iterations to 50 and evolve each skill for 10 steps with a batch size of 8. 

Following prior work \citep{yang2026skillopt}, we repeat the entire evolution process across three independent runs for each method to account for variability, and report the average test performance across the three resulting evolved skill sets. We split SpreadsheetBench-Verified using a fixed random seed of 42.

\textbf{LLM Skill.} LLM skill is directly generated by a single LLM call according to task descriptions. The generation prompt is "You are an expert in spreadsheet manipulation. Write a skill guide (Markdown) for an AI agent that solves Excel/xlsx tasks using Python (openpyxl and pandas). Be specific and practical."

\textbf{Human Skill.}
Following Trace2Skill \citep{ni2026trace2skill}, we adopt Anthropic's official XLSX skill as a human-written baseline, which provides expert-crafted guidelines for spreadsheet manipulation, including formula handling, data processing, and formatting. The skill is directly evaluated without further evolution.

\textbf{Trace2Skill.}
Trace2Skill \citep{ni2026trace2skill} employs a three-stage pipeline comprising trajectory generation, parallel patch proposal, and hierarchical consolidation. First, it collects successful and failed execution trajectories by evaluating the agent with the current skill on evolution tasks. Next, parallel success and error analysts extract reusable strategies from successful trajectories and diagnose the root causes of failures, respectively, to propose skill patches. Finally, a hierarchical merge operator consolidates these patches by eliminating redundancies and resolving conflicts, and applies the resulting updates to construct a skill document.

\textbf{EvoSkill.}
EvoSkill \citep{alzubi2026evoskill} frames skill evolution as an iterative search over a bounded frontier of candidate skills. At each iteration, it selects a parent skill from the frontier in a round-robin manner and evaluates it on sampled training tasks. The proposer analyzes failed execution trajectories and a cumulative history of previous proposals to generate candidate skill modifications. These modifications are applied to \texttt{SKILL.md} files, and the resulting candidates are evaluated on a held-out validation set. Candidates are retained in the frontier only if they outperform its weakest member.

\subsection{Analysis of Cross-model Transferability}
Table~\ref{tab:cross_model_transfer} reports all results of \method{} across three different model users and creators. We observe two clear trends. First, stronger skill users generally make better use of evolving skills. Averaged across our three skill creators, Qwen3.7-Plus achieves the highest overall performance than Qwen3.6-27B and Gemma4-31B, suggesting that a stronger inference model can more effectively interpret and execute the procedural knowledge encoded in the evolving skills. Notably, the skills generated by Gemma4-31B are executed more effectively by the other two models, even outperforming their respective self-evolved skills. Second, skills evolved by stronger creator models exhibit better transferability. For all three skill users, skills produced by Qwen3.7-Plus achieve the highest average performance. when transferred to Gemma4-31B and Qwen3.6-27B, the skills from Qwen3.7-Plus achieve average scores of 64.09 and 60.58 performance. Overall, these results indicate that the effectiveness of cross-model skill transfer depends on both the capability of the model that produces the skill and that of the model that uses it, while the evolved skills remain broadly reusable across heterogeneous backbone models.

\subsection{Prompt Templates Used in \method{}}
As shown in Figure~\ref{fig:inference_prompt}, we present the prompt templates used to solve spreadsheet tasks. The prompt instructs the inference agent to iteratively interactive with input spreadsheet, use bash tool to execute python code to read or modify the spreadsheet, and generate the output spreadsheet. We employ consistent prompt templates across all methods to ensure fair comparison and reproducibility.

As shown in Figure~\ref{fig:proposer_prompt}, we present the prompt templates used to propose an update patch. The prompt instructs the proposer to use predefined programmatic tools to read relevant information, including trajectory steps, the current skills with programs, and historical proposed patches on demond, and then propose the patch.

\subsection{Case Study}
\label{appendix:case_study}
In Table~\ref{tab:case_study}, we present a qualitative case study to illustrate how \method{} enables more precise skill evolution by separating descriptive knowledge from executable procedural logic. The task requires extracting two-digit month codes from spreadsheet strings, retrieving their corresponding month names from a reference table, and writing the results into designated cells. The text-only baseline, Trace2Skill, captures the general task-solving procedure but introduces potentially misleading rules during skill evolution. Specifically, it suggests allowing prefix matching when exact matches are unavailable and inferring month names directly from numeric codes when lookup fails. Although these rules appear reasonable, they may lead to incorrect matches or outputs inconsistent with the reference table, such as generating ``September'' instead of the expected ``SEP''. Furthermore, mixing these execution details with descriptive guidance makes it difficult to isolate and repair the underlying logic.

In contrast, \method{} separates the two forms of knowledge into a natural-language skill document and an associated executable program. The skill document preserves high-level task guidance and explicitly emphasizes the reference table as the authoritative source, while the program implements precise string extraction and exact lookup operations. Through programmatic patches, \method{} can directly refine the execution logic without modifying unrelated descriptive knowledge. Meanwhile, textual patches can independently improve task-level guidance without disrupting the program implementation. This case demonstrates that decoupling descriptive knowledge from procedural logic facilitates more targeted and reliable skill updates, reducing the risk of introducing ambiguous or conflicting execution rules during skill evolution.

\begin{figure*}[h]
\centering
\includegraphics[
    width=\linewidth,
    clip
]{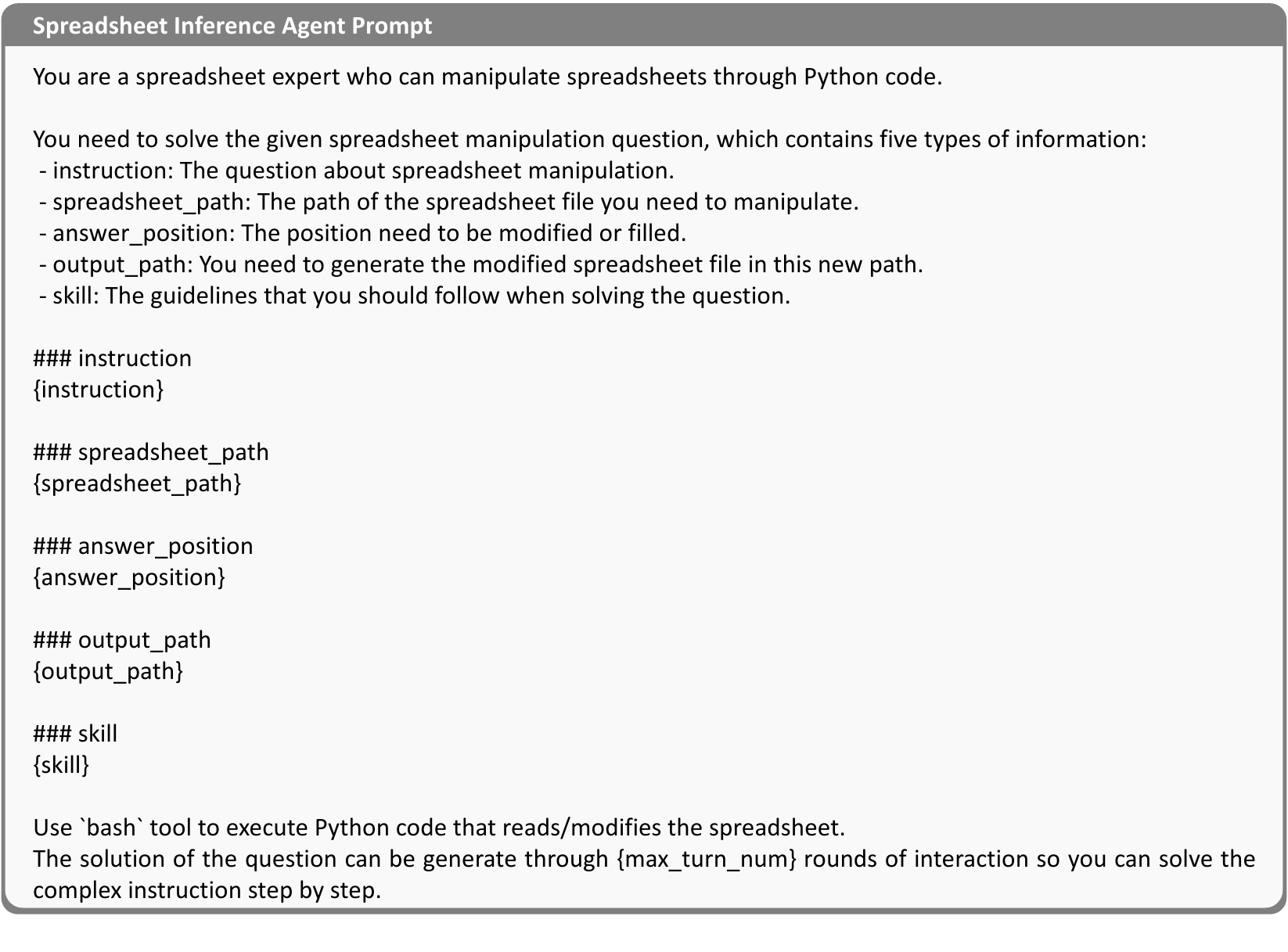}
\caption{Spreadsheet Inference Agent Prompt.}
\label{fig:inference_prompt}
\end{figure*}
\begin{figure*}[h]
\centering
\includegraphics[
    width=\linewidth,
    clip
]{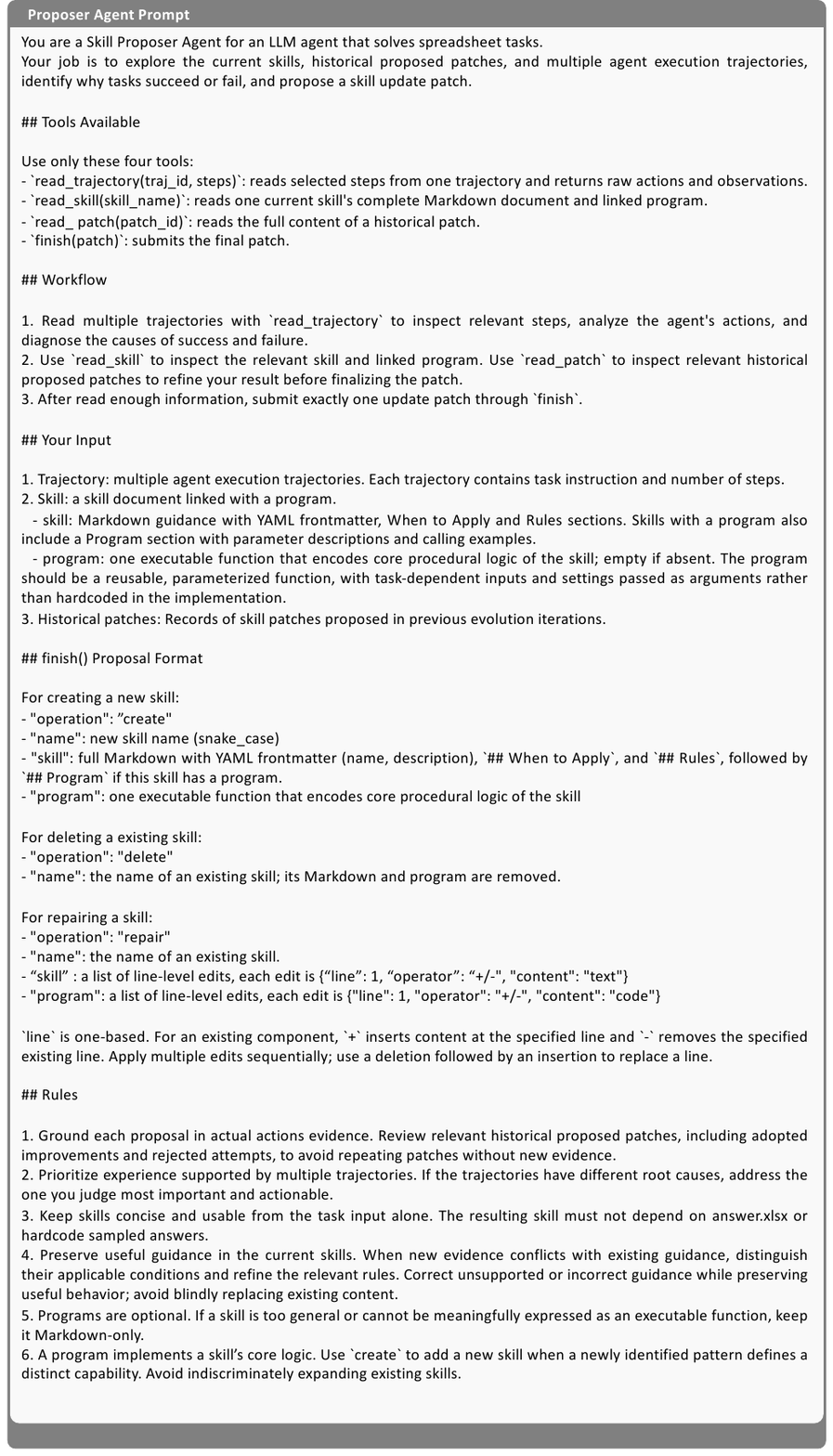}
\caption{Proposer Agent Prompt.}
\label{fig:proposer_prompt}
\end{figure*}
\begin{table*}[t]
\centering
\small
\renewcommand{\arraystretch}{1.5}
\resizebox{\linewidth}{!}{
\begin{tabular}{p{\linewidth}}
\toprule
\textit{\textbf{Case: Spreadsheet Manipulation: Extract and Look up}} \\
\midrule

\textbf{Instruction:}
For each string in cells A3:A14, extract the 2-digit month code and look up its corresponding month name in columns D:E. Fill cells B3:B14 with the results.
\quad\quad
\textbf{Ground Truth Answer:} B3 = SEP, B4 = JAN, B5 = DEC, ... \\

\hdashline

\textbf{Trace2Skill (Text-only Skill):} \\

name: Month Code Lookup \\

description: Extract month codes from spreadsheet strings and map them to month names using a lookup table. \\

\#\# When to Apply \\

Use when spreadsheet cells contain embedded month codes that need to be mapped to corresponding month names using a reference table. \\

\#\# Rules \\

1. Inspect the input strings to identify the location of the 2-digit month code. \\

2. Extract the month code using the appropriate string delimiter (e.g., the second token after splitting by "\_"). Preserve leading zeros (e.g., "09"). \\

3. Keep extracted month codes as strings during lookup. \bad{Allow prefix matching when exact matches are unavailable to accommodate formatting variations.} \\

4. Locate the month lookup table and retrieve the month name corresponding to the extracted code. \bad{If the lookup fails, infer the month name from the numeric code as a fallback.} \\

5. Write each matched month name into the corresponding target cell, preserving the original row alignment. \\

\hdashline

\textbf{Inference of Trace2Skill:} \\

{\ttfamily
from openpyxl import load\_workbook; import calendar;
wb=load\_workbook("input.xlsx"); ws=wb.active;
m=\{str(ws.cell(r,4).value).strip():ws.cell(r,5).value
for r in range(3,ws.max\_row+1)
if ws.cell(r,4).value is not None and ws.cell(r,5).value is not None\};
lookup=lambda key: m.get(key) or
\bad{next((v for k,v in m.items() if k.startswith(key)),None)}
or \bad{(calendar.month\_name[int(key)] if key.isdigit()
and int(key) in range(1,13) else None)};
[ws.cell(r,2,lookup(str(ws.cell(r,1).value).split("\_")[1].strip()))
for r in range(3,15)
if ws.cell(r,1).value is not None and "\_" in str(ws.cell(r,1).value)];
wb.save("output.xlsx")
} \\

\hdashline

\textbf{\method{} (Skill):} \\

name: Month Code Lookup \\

description: Retrieve month names from spreadsheet strings using a reference lookup table. \\

\#\# When to Apply \\

Use when spreadsheet cells contain embedded month codes that need to be mapped to corresponding month names using a reference table. \\

\#\# Rules \\

1. Inspect the input strings to identify the location of the 2-digit month code. \\

2. \good{Use the linked program to extract month codes and perform exact lookup.} \\

3. \good{Treat the reference table as the authoritative source and avoid inferring month names when no matching entry exists.} \\

4. Write each matched month name into the corresponding target cell, preserving the original row alignment. \\

\#\# Program \\

Parameters: \texttt{lookup\_month(token, m)}, where \texttt{token} is the input string containing a month code, and \texttt{m} is a dictionary mapping month codes to month names. \\

Usage Example: \texttt{lookup\_month("demo\_07\_record", \{"07": "July"\})} $\rightarrow$ \texttt{"July"} \\

\hdashline

\textbf{\method{} (Program):} \\

{\ttfamily
import re;
def lookup\_month(token, m):
\good{key = re.split(r"[-\_]", token)[1]};
\good{value = m.get(key, None)};
return value
} \\

\hdashline

\textbf{Inference of \method{}:} \\

{\ttfamily
from openpyxl import load\_workbook;
wb=load\_workbook("input.xlsx"); ws=wb.active;
m=\{str(ws.cell(r,4).value).strip():ws.cell(r,5).value
for r in range(3,ws.max\_row+1)
if ws.cell(r,4).value is not None and ws.cell(r,5).value is not None\};
[ws.cell(r,2,\good{lookup\_month(str(ws.cell(r,1).value),m)})
for r in range(3,15)
if ws.cell(r,1).value is not None
and any(d in str(ws.cell(r,1).value) for d in ("-", "\_"))];
wb.save("output.xlsx")
} \\

\bottomrule
\end{tabular}
}
\caption{
Illustrative case study comparing text-only skill evolution with
\method{} on spreadsheet manipulation.
Potentially incorrect knowledge and operations are highlighted in
\bad{red}, while correctly refined textual knowledge and program logic
are highlighted in \good{blue}.
}
\label{tab:case_study}
\end{table*}

\end{document}